\pdfoutput=1

\documentclass[11pt]{article}
\usepackage{ACL2023}
\usepackage{graphicx}
\usepackage{booktabs} 
\usepackage{threeparttable} 
\usepackage{multirow}
\usepackage{amssymb} 
\usepackage{bm}
\usepackage{array}
\usepackage{times}
\usepackage{makecell}
\usepackage{latexsym}
\usepackage{algorithm}
\usepackage{graphicx} 
\usepackage{algorithm}
\usepackage{algpseudocode}
\usepackage{amsmath} 
\usepackage[utf8]{inputenc}
\usepackage{graphicx}
\usepackage{booktabs}
\usepackage{tabularx}
\usepackage{amssymb}
\usepackage{xcolor}
\usepackage{xspace}
\usepackage{subcaption}

\usepackage[T1]{fontenc}

\usepackage[utf8]{inputenc}

\usepackage{microtype}
\usepackage{inconsolata}

\usepackage{hyperref}

\newcommand{\eg}{\emph{e.g.},\xspace}
\newcommand{\ie}{\emph{i.e.},\xspace}

\newcommand\figref[1]{Fig.~\ref{#1}}

\newcommand\tabref[1]{Table~\ref{#1}}

\newcommand\secref[1]{Sec.~\ref{#1}}

\newcommand{\fakeparagraph}[1]{\vspace{1mm}\noindent\textbf{#1.}}

\newcommand{\sysname}{StablePT\xspace }

\title{\sysname: Towards Stable Prompting for Few-shot Learning via Input Separation}
\author{Xiaoming Liu\textsuperscript{$1$ $\dagger$ *}, Chen Liu\textsuperscript{$1$ $\dagger$}, Zhaohan Zhang\textsuperscript{$2$}, Chengzhengxu Li\textsuperscript{$1$}, \\\textbf{Longtian Wang\textsuperscript{$1$},  Yu Lan\textsuperscript{$1$},  Chao Shen\textsuperscript{$1$}} \\
        \textsuperscript{1}Faculty of Electronic and Information Engineering, Xi'an Jiaotong University, Xi'an, China\\
        \textsuperscript{2}Queen Mary University of London, London, UK \\
        \textsuperscript{$\dagger$} Eqal contribution, \textsuperscript{*} Corresponding author\\
        \texttt{
        \{xm.liu,ylan2020,chaoshen\}@xjtu.edu.cn}
        \\
        \texttt{\{lcoder,cxz.li,w976625943\}@stu.xjtu.edu.cn},
        \texttt{zhaohan.zhang@qmul.ac.uk}
        }

\begin{document}
\maketitle
\begin{abstract}
Large language models have shown their ability to become effective few-shot learners with prompting, revolutionizing the paradigm of learning with data scarcity.
However, this approach largely depends on the quality of prompt initialization, and always exhibits large variability among different runs.
Such property makes prompt tuning highly unreliable and vulnerable to poorly constructed prompts, which limits its extension to more real-world applications.
To tackle this issue, we propose to treat the hard prompt and soft prompt as separate inputs to mitigate noise brought by the prompt initialization.
Furthermore, we optimize soft prompts with contrastive learning for utilizing class-aware information in the training process to maintain model performance.
Experimental results demonstrate that \sysname outperforms state-of-the-art methods by 6.97\% in accuracy and reduces the standard deviation by 1.92 on average.
Furthermore, extensive experiments underscore its robustness and stability across 8 datasets covering various tasks.
\footnote{Codes are available at \href{https://github.com/lccc0528/Stable/tree/main}{https://github.com/lccc0528}.}

\end{abstract}

\section{Introduction}

Pre-trained Language Models (PLMs) \cite{ ouyang2022training, touvron2023llama, anil2023palm, OpenAI2023GPT4TR} exhibit an overwhelming capacity to understand, analyze and classify textual information.
Recent works suggest prompt tuning \cite{shin2020autoprompt, schick2021exploiting, ye2022ontology, han2022ptr, liu2023prompt, wu2024infoprompt} to be a plausible method for efficiently adapting abundant but general knowledge behind PLMs to various downstream tasks.
Prompt tuning reformulates downstream tasks into the same Mask Language Modeling (MLM) problem as pre-training of PLMs by constructing proper templates with open slots.
In this way, the tasks benefit from PLMs' generative capability to fill the slots.
And \citet{brown2020language} find that PLMs are able to learn well from limited training templates for certain tasks. 
Thus, it has been a practical topic to generalize PLMs' ability in a prompt manner with only a few samples for both training and data efficiency.
\begin{figure}
    \centering
    \includegraphics[width=\columnwidth]{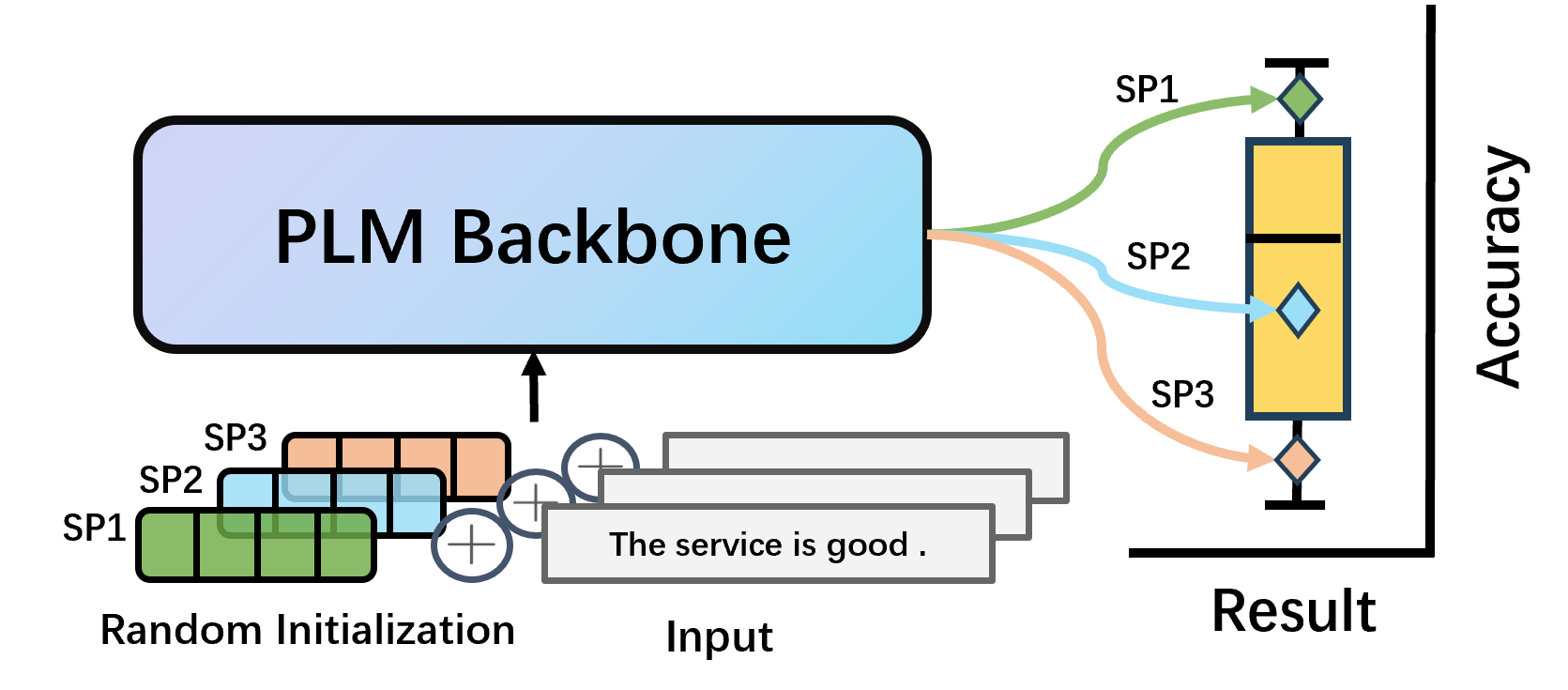}
    \includegraphics[width=0.48\columnwidth]{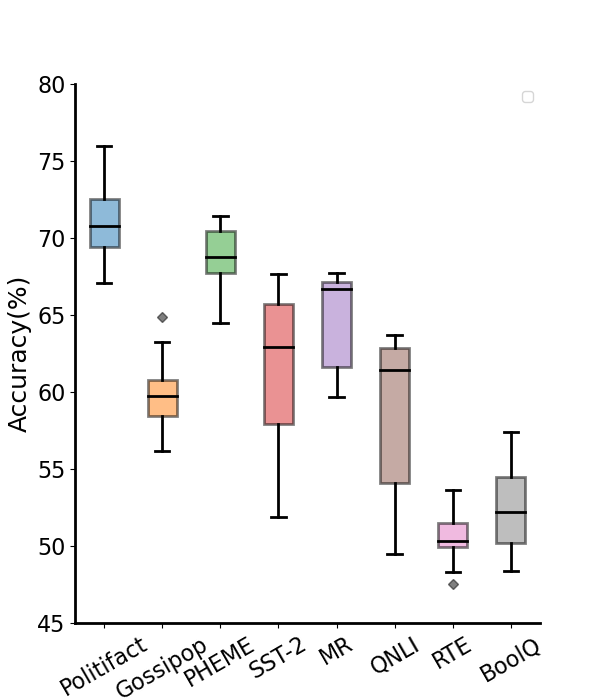}
    \includegraphics[width=0.48\columnwidth]{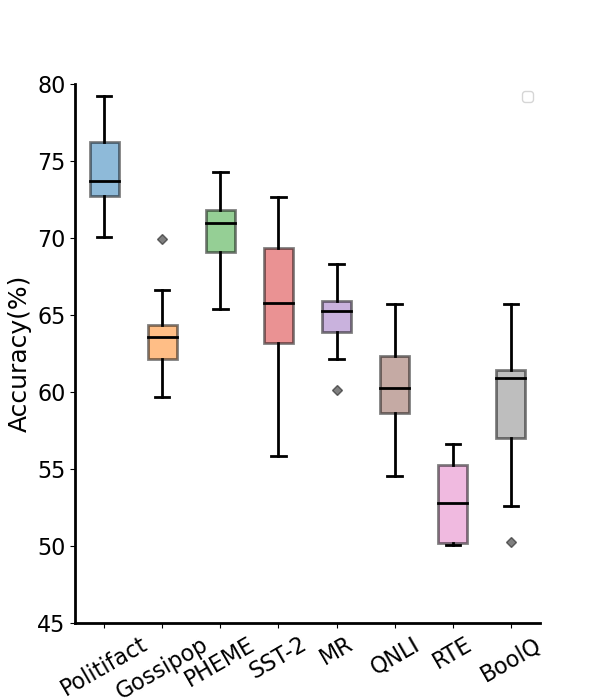}
    \caption{Illustration of performance various on NLU tasks with different prompt initialization. SP means soft prompt. The left subplot takes RoBERTa-base as the backbone, and the right shows RoBERTa-large.
    The accuracy variation can reach 15.77\% and 16.86\% on the SST-2 dataset, respectively.}
    \label{fig:intro1}
\end{figure}

Current works on the construction of prompts could be roughly categorized into two branches: 
a) \textbf{Hard prompt construction}, which generates a template in natural language manually \cite{petroni2019language, schick2021exploiting} or automatically \cite{jiang-etal-2021-know, ben2021pada, li2023dialogue}.
While hard prompts provide explicit guidance for PLMs, the performance is largely dependent on the selection of a proper template.
In fact, \citet{zhao2021calibrate} report that different prompt formats could lead to accuracy varying from near chance to near state-of-the-art.
b) \textbf{Soft prompt construction}, which searches
for appropriate prompts in embedding space \cite{li2021prefix, lester2021power}.
Although steering away from natural language avoids meticulous selection of manual prompts, effective initialization of soft prompts is considered crucial for gaining satisfying performance \cite{gu2022ppt}.
\figref{fig:intro1} illustrates the instability of vanilla prompt tuning with 10 runs on eight natural language understanding (NLU) tasks.
A common solution for searching valid prompt initialization in continuous space is pre-training \cite{gu2022ppt, vu2022spot}, 
which first learns the prompt on multiple datasets with diversified tasks but requires massive computing expenses.

To deal with the defects of hard and soft prompt construction, we propose a Stable Prompt Tuning method, named \sysname, which is robust to prompt initialization quality and keeps out performance and stability in the meantime.
Instead of concatenating hard and soft prompts directly, we treat them as separate inputs to different transformers-style modules. 
Hard prompts with contexts are embedded by PLM and a single SemEncoder layer for context-aware representation, and a GenDecoder layer takes soft prompt as input, further injecting class-aware information into cloze templates by contrastive learning.
Moreover, we apply supervised contrastive loss in the soft prompt optimization process, enhancing the model's ability to achieve inter-class separation and inner-class compactness, which further boosts model performance in the few-shot setting.
\textbf{The advantages of our method are three-folded:}
(\textit{i}) Reduce noise caused by undesired soft prompt initialization through processing hard prompt and soft prompt with different modules.
(\textit{ii}) Soft prompts provide the verbalizer with additional class-aware information, ensuring consistent performance regardless of the discrete template expression.
(\textit{iii}) Hard prompts offer explicit guidance on soft prompt optimization for extracting task-specific knowledge efficiently.

The contribution is summarized as follows:
\begin{itemize}
    \item \textbf{Input Separation:} We design a novel strategy that separates soft prompt from textual input to alleviate performance inconsistency brought by the initialization quality of continuous templates.
    \item \textbf{Information Fusion:} We design an interaction learning process for hard and soft prompt optimization, which integrates context-aware and class-aware information for stable performance.
    \item \textbf{Outstanding Performance:} \sysname surpasses state-of-the-art methods on 8 NLU tasks. The experiment results verify its robustness and stability across different prompt initializations.
\end{itemize}

\section{Related Work}

\fakeparagraph{Prompt Tuning}
Prompt tuning is an efficient approach to adapting PLMs to downstream tasks.
The initial method in prompt tuning \cite{schick2021exploiting,schick2021s} involves manually designing hard prompts composed of discrete words. 
Subsequent works \cite{gao2021making,jiang2020can,shin2020autoprompt} propose the automatic generation of prompts for mining appropriate templates to obtain desired outputs. 
Yet, previous work reveals that changing a single word in the hard prompt might result in a substantial performance drop \cite{liu2023gpt}.
Soft prompting is another branch of prompt tuning, which searches proper prompt in continuous vector space \cite{li2021prefix, qin-eisner-2021-learning, lester-etal-2021-power}. 
It endows model flexibility to various downstream tasks while suffering greatly from undesired initialization 
\cite{gu2022ppt}.
To keep consistent performance for prompt tuning, PPT \cite{gu2022ppt} pre-trains prompts with 10 GB textual data, and SPoT \cite{vu2022spot} conducts prompt pre-training on three tasks across eight datasets to attain transferable prompts.
However, prompt pre-training also requires massive computing expenses (\eg it takes 25 hours to pre-train PPT and 30 hours to pre-train SPoT with Roberta-base on a single NVIDIA A100), which is contrary to the original intention of prompt tuning.
Our method achieves the goal of stabilizing language model adaptation, \ie reducing the model performance fluctuates for different prompt initialization \cite{liu2023gpt}, by disentangling hard prompt and soft prompt to avoid noise propagation, refraining the necessity to pre-train prompts.

\fakeparagraph{Few-shot Learning with PLMs} 
Prompting PLMs, such as GPT-3 \cite{brown2020language} and PET \cite{schick2021exploiting}, are found to obtain surprising performance in substantial downstream tasks with few-shot settings.
Subsequent research efforts by \citet{perez2021true} and \citet{bragg2021flex} have explored reasonable few-shot settings by constraining the size of the validation set and introducing a unified framework for assessing few-shot performance.
The study by \citet{logan2022cutting} emphasizes that fine-tuning PLMs in few-shot learning scenarios can enhance the model's robustness to different hard prompts.
However, their approaches do not address the fundamental issue of poor prompt adjustment performance in few-shot scenarios. 
Our method overcomes this by fine-tuning trivial portions in additional layers with interaction between disentangled prompts.

\section{Methodology}

\begin{figure*}
    \centering
    \includegraphics[width=\linewidth]{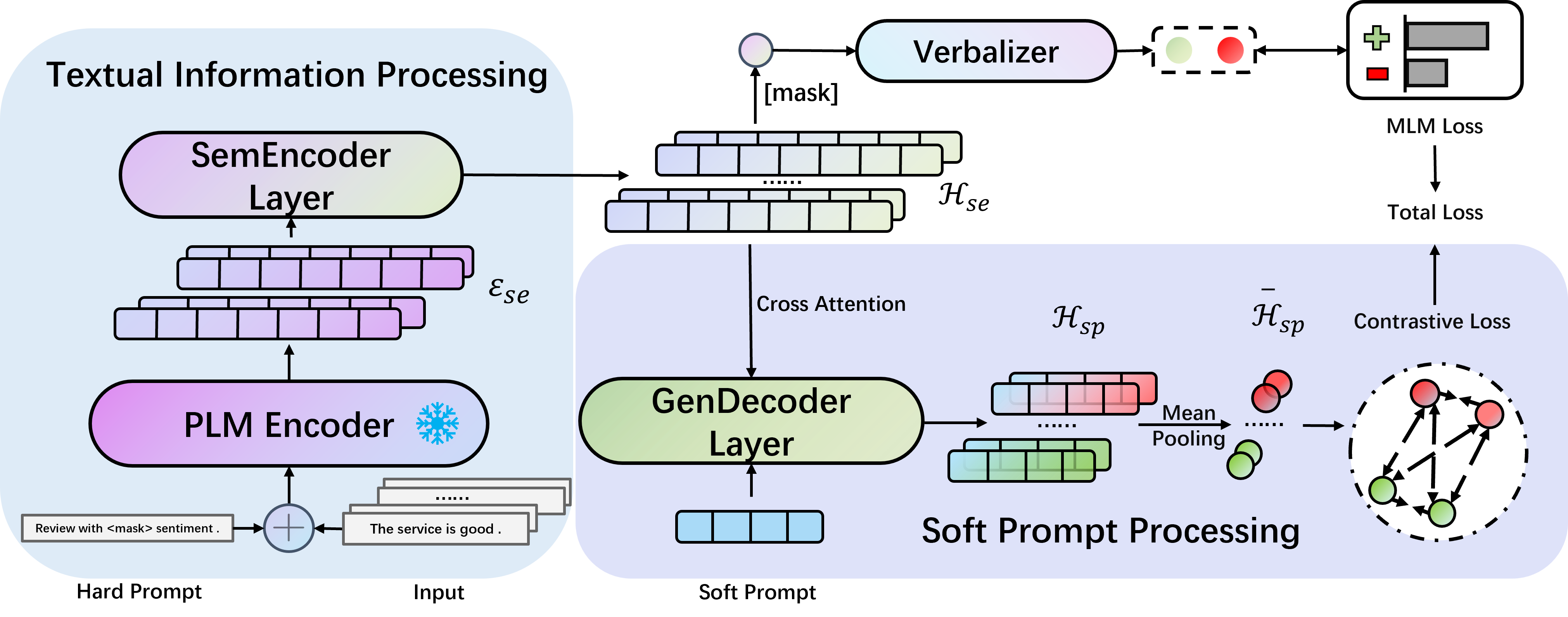}
    \caption{Overview of \sysname. 
    The textual information processing, through a dual-encoding module, coordinates task-specific invariant prompts, simplifies adaptation to various language comprehension tasks, and enhances the semantic processing capabilities of static PLMs through dynamic semantic encoding (\secref{sec:hardppt}). 
    The generative decoder module is central to the model, processing soft prompts through a cross-attention mechanism to interact with the encoder's outputs and diminish the adverse effects of random initialization (\secref{sec:GD}). 
    The training regime employs a dual loss function comprising MLM loss and contrastive loss, aiming at enhancing the model's capabilities in language comprehension and categorical differentiation (\secref{sec:MT}). 
      }  

    \label{fig:framework}
\end{figure*}

We introduce a novel language model architecture named \sysname that processes textual information\footnote{Textual information refers to the information formed by concatenating the hard prompt and the input.} and soft prompts separately but keeps interaction between them.
It helps stabilize model performance across different initialization of hard/soft prompts.
The detailed illustration of \sysname is shown in \figref{fig:framework}.

\subsection{Textual Information Processing}
\label{sec:hardppt}
To exploit the strengths of hard prompts in few-shot learning while addressing their robustness issues, we introduce a prompt-based multi-encoder architecture to process textual information. This architecture encapsulates a dual-encoding module for context-aware information extraction and class-aware information absorption.

\fakeparagraph{Task-Specific Hard Prompt} 
Task-specific hard prompts can provide clear context-relevant guidance for the model, helping PLMs accurately focus on task-related semantic representations.
For example, in sentiment analysis, the hard prompt \textit{"the sentiment of the review is [mask]"} directly guides the model to focus on the emotional polarity of the review text, thereby improving the prediction accuracy of a specific task.

\fakeparagraph{PLM Encoder} The PLM Encoder is the cornerstone of the architecture, harvesting the extensive knowledge accumulated during the PLM's pre-training. 
It anchors the initial processing of the hard prompt and input, encoding them into semantic embeddings.
We set PLM in a frozen state according to the general prompt tuning setting.
Despite being in a frozen state, the PLM Encoder retains its efficacy in capturing the core semantics intended by the hard prompts.
Specifically, the original hard prompt template $\mathcal{P}_{h}$ and the input sentence $x$ are syntactically mapped into a semantic vector space:

\begin{equation}
    \mathcal{E}_{se}=\text{PLMEncoder}([\mathcal{P}_{h},x])
\end{equation}
where $[\cdot,\cdot]$ is the splicing operation, $\mathcal{E}_{se} \in \mathbf{R}^{b \times o \times d}$ is the embeddings of the hard prompt and input sentence, $b$ is batch size, $o$ is the maximum sequence length of PLMs, and $d$ is the embedding dimension of PLMs.

\fakeparagraph{Semantic Encoder Layer} 
Building on the foundation of the PLM encoder, we initialize a single trainable transformer encoder layer, which adeptly projects hard prompts and inputs into a shared space by self-attention mechanism while incorporating class-aware information back-propagated from soft prompt processing module, ensuring that semantically similar embeddings are projected more consistently within the space.
Specifically, the semantic encoder refines and recontextualizes the embeddings $\mathcal{E}_{se}$ produced by the PLM encoder, further modeling semantic interactions:
\begin{gather}
\mathcal{H}_{se}=\text{softmax}(\frac{Q_EK_E^T}{\sqrt{d_k}})V_E
\end{gather}
where $Q_E$, $K_E$, $V_E$ are linear transformation of $\mathcal{E}_{se}$ with matrix $W^E_q$, $W^E_k$, $W^E_v$ which are initialized in semantic encoder layer.

Overall, the prompt-based multi-encoder maps discrete prompts and inputs into a syntactic embedding space guided by the prior knowledge of the PLM encoder. These embeddings are then intricately processed within the semantic encoder for semantic interaction. Subsequently, at the top of the semantic encoder, a verbalizer is used to map the output states $\mathcal{H}^{m}_{se}$ of the [\textit{mask}] token to the label $y$. Furthermore, the output states $\mathcal{H}_{se}$ is fed into a generative decoder to interact with soft prompts, thus completing the semantically enriched cycle of the architecture.
\subsection{Soft Prompt Processing}
\label{sec:GD}
In order to mitigate the issue of substantial noise caused by the random initialization of soft prompts in few-shot scenarios, we introduce a transformer decoder layer called GenDecoder Layer for soft prompts to interact with semantic output states $\mathcal{H}_{se}$ instead of concatenating soft prompts on the input textual information.
We initialize a soft prompt of length $l$ with random values. 
This soft prompt serves as a query input to the GenDecoder Layer, while the output states from the SemEncoder Layer are used as key and value inputs to the decoder. 
Through the cross-attention mechanism, the soft prompt functions as a query to retrieve information from the hidden states, thereby achieving the objective of interacting with the input and extracting context-aware information.
Specifically, we apply cross-attention with $\mathcal{P}_{s}$ as query and $\mathcal{H}_{se}$ as key and value:
\begin{gather}
  Q_D =  W^D_q\mathcal{P}_{s}, \notag \\
  K_D,V_D = W^D_k\mathcal{H}_{se},W^D_v\mathcal{H}_{se},\notag \\
 \mathcal{H}_{sp}=\text{softmax}(\frac{Q_DK_D^T}{\sqrt{d_k}})V_D
\end{gather}
where $\mathcal{H}_{sp}\in\mathbf{R}^{b \times l \times d}$ is the output states of soft prompt, $W^D_q, W^D_k, W^D_v$ are initilzed in GenDecoder Layer.
$\mathcal{H}_{sp}$ is further processed through mean pooling to obtain more representative embeddings:
\begin{equation}
\bar{\mathcal{H}_{sp}}=\text{Meanpooling}([\mathcal{H}_{sp}])
\end{equation}

Furthermore, we apply a supervised contrastive learning method \cite{gunel2020supervised} for extracting class-aware information from $\bar{\mathcal{H}_{sp}}$.
Our intention is to achieve instance-level intra-class compactness and inter-class separability for discovering essential differences between classes.
The contrastive learning loss is formulated as:

\begin{small}
\begin{align}
\mathcal{L}_{C L} =& -\frac{1}{b} \sum_{j=1}^b \bm{1}_{\left[y_i=y_j\right]} \nonumber \\
& \log \frac{\exp \left(\operatorname{sim}\left(\bar{\mathcal{H}^{(i)}_{sp}},  \bar{\mathcal{H}^{(j)}_{sp}}\right) / \tau\right)}{\sum_{K=1}^B \exp \left(\operatorname{sim}\left(\bar{\mathcal{H}^{(i)}_{sp}}, \bar{\mathcal{H}^{(k)}_{sp}}\right) / \tau\right)}
\label{eq5}
\end{align}
\end{small}
where $\bm{1}$ is the indicator, $\operatorname{sim}(\cdot,\cdot)$ denotes the normalized cosine similarity score, and $\tau$ is the temperature coefficient.

Soft prompt processing aligns soft prompts more closely with the semantic representation encoded by the encoder and injects class-aware information obtained from contrastive learning into context-aware representation.
In this way, model performance is stabilized regardless of prompt initialization because changing a single word in the hard prompt would not lead to insufficient class-aware information extraction, and random soft prompt initialization would not affect the textual information processing stage.

\subsection{Loss Function Design}
\label{sec:MT}
We implement Masked Language Model (MLM) loss for filling slots in prompts, which is derived from the output states of the encoder. Specifically, for the masked token representations $\mathcal{H}^{m}_{se}$, a verbalizer function is employed to project these representations into a space where the MLM loss can be calculated. 
\begin{align}
&\mathcal{L}_{MLM}=\arg\max_\theta \sum_x \log p(y|[\mathcal{P}_{h},x]; \theta) \nonumber \\
&= \arg\max_\theta \sum_x \log p(\langle X \rangle = v(y)|\mathcal{H}^{m}_{se} ; \theta)
\label{eq6}
\end{align}
where $y$ is the label of input sentence $x$, $\langle X \rangle$ is the mask token, $v(\cdot)$ is the verbalizer of PLMs, $\theta$ indicates all tunable parameters. 
The total loss for the model training is the sum of these two components, aiming to optimize both the language understanding and categorical distinction capabilities of the model.
\begin{equation}
\mathcal{L}_{total}= \mathcal{L}_{MLM} + \mathcal{L}_{CL}
\label{eq7}
\end{equation}
The pseudocode of the training process is shown in Appendix \ref{pscode}.

\section{Experiments}

\subsection{Datasets and Settings}

We evaluate our model across eight datasets on four NLU tasks: \textbf{Fake News Detection} including Politifact~\cite{shu2020fakenewsnet}, Gossipop~\cite{shu2020fakenewsnet} and PHEME~\cite{buntain2017automatically}; \textbf{Sentiment Analysis} including SST-2~\cite{socher2013recursive} and MR~\cite{pang2005seeing}; \textbf{Natural Language Inference (NLI)} including QNLI~\cite{wang2018glue} and RTE~\cite{wang2018glue}; \textbf{Question Answer (QA)} including BoolQ~\cite{wang2019superglue}.
The experiment details and the impact of hyperparameters are shown in Appendix \ref{DatasetInfo} and Appendix \ref{hyperparameter}, respectively.
\begin{table*}[h]
    \renewcommand\arraystretch{1.2}
    \centering
    \small
    \begin{tabularx}{\textwidth}{p{1.88cm}| >{\centering\arraybackslash}m{1.05cm}>{\centering\arraybackslash}m{1.05cm}>{\centering\arraybackslash}m{1.25cm}|>{\centering\arraybackslash}m{1.05cm}>{\centering\arraybackslash}m{1.25cm}|>{\centering\arraybackslash}m{1.05cm}>{\centering\arraybackslash}m{1.25cm}|>{\centering\arraybackslash}m{1.05cm} c}
    \toprule
        \textbf{Method} & \textbf{Politifact} & \textbf{Gossipop} & \textbf{PHEME} & \textbf{SST-2} & \textbf{MR} & \textbf{QNLI} & \textbf{RTE}& \textbf{BoolQ}&\textbf{Avg.}\\ 
    \midrule
        Prompt Tuning & $70.88_{2.54}$ & $59.10_{2.72}$ & $67.84_{1.96}$ & $62.31_{5.56}$ & $64.75_{3.03}$ &$59.34_{5.07}$&$50.51_{1.84}$ &$52.03_{3.04}$ &$60.85$ \\ 
        Prefix Tuning & $72.91_{3.71}$& $58.77_{2.31}$ & $68.73_{2.17}$ & $70.31_{1.91}$ & $67.84_{2.62}$  & $60.27_{2.33}$ & $50.82_{1.71}$ & $54.32_{2.79}$ &$63.00$ \\ 
        P-Tuning V2 & $70.99_{6.69}$ & $59.99_{4.84}$ & $72.31_{2.16}$ & $68.71_{3.67}$ & $68.27_{1.34}$ & $60.73_{2.44}$ &$51.16_{0.73}$ &$58.64_{1.77}$&$63.85$ \\ 
        PPT  & $73.00_{2.45}$ & $61.58_{1.42}$ & $69.55_{1.23}$ & $73.83_{0.64}$ & $69.92_{1.26}$ & $60.40_{1.46}$ & $50.62_{1.32}$ &$55.33_{1.08}$ &$64.28$ \\ 
        SPoT & $75.43_{1.94}$ & $65.18_{1.07}$ & $72.55_{1.21}$ & $74.79_{0.82}$ & $70.39_{0.77}$ & $61.49_{1.30}$& $51.38_{1.09}$ &$56.34_{1.17}$ &$65.94$\\ 
        SMoP & $73.16_{2.83}$ & $61.45_{2.98}$ & $70.80_{3.62}$ & $70.67_{3.05}$ & $68.19_{3.57}$ & $60.33_{1.60}$ & $50.96_{0.58}$ &$55.29_{2.12}$&$63.86$\\ 
        E$^{2}$VPT & $78.62_{1.66}$ & $68.32_{2.49}$ & $72.16_{1.27}$ & $70.35_{1.96}$ & $67.75_{4.54}$ & $60.78_{2.07}$ & $53.77_{2.06}$ &$57.93_{2.47}$&$66.21$\\ 
        ResPT & $78.46_{1.48}$ & $67.05_{1.69}$ & $71.45_{2.26}$ & $75.92_{3.11}$ & $71.35_{2.73}$ & $62.02_{1.69}$ & $54.57_{2.10}$
        &$59.54_{2.07}$&$67.55$ \\ 
        
    \midrule
        Fine Tuning &$79.38_{2.63}$&$67.31_{2.45}$&$75.01_{2.26}$ &$80.07_{3.92}$& $79.52_{1.93}$& $62.83_{4.37}$ &$52.21_{1.82}$ &$61.69_{1.58}$&$69.75$\\
        CP-Tuning &$79.52_{2.28}$&$69.80_{2.83}$&$74.29_{2.52}$&$81.00_{4.31}$& 
        $80.79_{1.21}$ & $63.02_{1.43}$ &$55.83_{4.15}$
        &$61.12_{1.67}$&$70.67$ \\
    \midrule
        \sysname & $\bm{80.86_{0.83}}$ & $\bm{72.16_{0.72}}$ & $\bm{75.35_{0.81}}$ & $\bm{84.04_{0.45}}$ & $\bm{81.84_{0.57}}$ & $\bm{63.37_{0.91}}$& $\bm{60.34_{0.55}}$ &$\bm{62.54_{0.78}}$&
        $\bm{71.31}$\\
    \bottomrule
    \end{tabularx}
    \caption{Comparison of \sysname and baseline methods on few-shot NLU tasks in accuracy. 
    The subscript means the standard deviation (\eg $80.86_{0.83}$  means 80.86±0.83) and the same to the following Tables. 
    }
    \label{main_exp}
\end{table*}

\subsection{Comparison Methods}
We conduct extensive experiments with 10 competitors that are reported to be effective in few-shot settings.
The comparison methods are divided into two categories roughly, \ie \textbf{prompt-tuning} and \textbf{fine-tuning}.  

\fakeparagraph{Prompt Tuning~\cite{lester-etal-2021-power}} It prepends a sequence of soft prompt tokens to the input and only tunes the soft prompt for adaptation. 

\fakeparagraph{Prefix Tuning~\cite{li2021prefix}} It reparameterizes networks for soft prompts and integrates and adjusts soft prompts at every layer of the PLM. 

\fakeparagraph{P-Tuning v2~\cite{liu2021p}} It can opt for a reparameterization network, and utilize the [CLS] classification head to adjust the soft prompts at each layer of the PLM. 

\fakeparagraph{PPT~\cite{gu2022ppt}} It uses designed pattern-verbalizer pairs on large-scale unlabeled data for self-supervised learning to pre-train soft prompts. 

\fakeparagraph{SPoT~\cite{vu-etal-2022-spot}} It pre-trains soft prompts on multi-task datasets through transfer learning and investigates the transferability between tasks. 

\fakeparagraph{SMoP~\cite{choi2023smop}} It employs a gating mechanism to use multiple short soft prompts specific to data subsets as an alternative to tuning with a single long soft prompt. 

\fakeparagraph{E$^{2}$VPT~\cite{han20232vpt}} It is a visual prompting method but could generalize well to NLU task. It introduces key-value prompts in the self-attention module of PLM and jointly serves as learnable parameters with soft prompts from the input layer in the model. 

\fakeparagraph{ResPT~\cite{razdaibiedina2023residual}} It employs a residual connection network as a reparameterization network to adjust the soft prompts of the input layer.

\fakeparagraph{Fine Tuning~\cite{kenton2019bert}} It fine-tunes the model for classification by adding a linear classification layer on top of PLMs. 

\fakeparagraph{CP-Tuning~\cite{xu2023making}} It adds soft prompts to the end of the input to provide a novel contrastive loss and combines with an additional MLM loss for prompt-oriented fine-tuning. 

For a fair comparison, we reimplement PPT and SPoT with the same backbone model, \ie Roberta-base. We pre-train SPoT on GLUE datasets~\cite{wang2018glue}, which enables SPoT to achieve the best results reported in original paper. Moreover, to conduct a fair comparison, we exclude SST-2 and RTE. We pre-train PPT on the same pre-training dataset as the original paper, \ie 10 GB OpenWebText \cite{gokaslan2019openwebtext}. Further model-level comparisons are shown in Appendix \ref{methods}.

\subsection{Comparison Results}
\label{sec:main}
The comparison results on eight datasets are listed in \tabref{main_exp}. Overall, \sysname outperforms all baselines by a large margin on all three different tasks and keeps the lowest standard deviation.

From the results, we have the following key findings: 
(\textit{i}) \sysname demonstrates superior performance on all datasets, even surpassing the methods that involve full-model fine-tuning (\ie Fine Tuning, CP-Tuning) by at least 1.34\%, 2.36\%, 3.04\% and 4.51\% on the Politifact, Gossipop, SST-2 and RTE datasets.
It benefits from our novel interaction mechanism which significantly mitigates the issues caused by prompt initialization in low-resource settings, even without pre-training of the prompts.
(\textit{ii}) Random initialization of prompts causes significant instability of results across all datasets.
But our method \sysname demonstrates superior stability with an average standard deviation of 0.7 across datasets, outperforming Prompt Tuning, Prefix Tuning, P-Tuning V2 and SMoP in standard deviations by 3.2, 2.4, 3.1, and 2.6 respectively.
(\textit{iii}) Prefix Tuning, P-Tuning V2, E$^{2}$VPT, and ResPT generally outperform vanilla prompts in most scenarios, demonstrating that the additional parameters effectively enhance model performance.
However, reparameterization can also exacerbate result instability in certain situations. 
For example, on Politifact, the standard deviations of Prefix Tuning and P-Tuning V2 are higher than Prompt Tuning by 1.2 and 4.2, respectively.
(\textit{iv}) Pre-trained prompts might lead to suboptimal performance due to inappropriate knowledge transfer caused by domain inconsistency between pre-training and downstream tasks. In contrast, multi-task pre-training tends to outperform single-task pre-training in downstream applications. For instance,  PPT and SPoT show a smaller performance gap in sentiment analysis at 0.7\% than in fake news detection at 3.0\%.

\subsection{Stability to Prompts Initialization}\label{HPI}
We would show the better stability of \sysname to various prompt initializations, which is compared with hybrid prompt tuning (Hybrid PT) \cite{liu2023gpt} on two different tasks.
\begin{table}[h!]
\small
\centering
\begin{tabularx}{\columnwidth}{>{\centering\arraybackslash}X >{\centering\arraybackslash}X c|>{\centering\arraybackslash}X c}
\toprule
& \multicolumn{2}{c|}{\textbf{SST-2}} &  \multicolumn{2}{c}{\textbf{PHEME}} \\ \midrule

Method & \sysname & Hybrid PT  & \sysname & Hybrid PT \\ \midrule
Random & $\underline{82.27}$&$71.15$& $76.22$ & \(70.45\) \\ 
Label & \(82.98\)& \(71.94\)& \(76.53\) & $\underline{65.09}$ \\ 
Vocab &\( 83.03 \)& $\bm{74.21}$ & $\underline{76.09}$      &\(72.75\) \\ 
Top-1k & \( 83.05 \)& $\underline{70.29}$ & \(76.35\)   &\(73.45\)     \\ 
Task & $\bm{83.10}$ & \(73.76\)  & $\bm{76.79}$ & $\bm{73.95}$ \\ 
\midrule
\textbf{Std.}   &\(0.31\)&\(1.50\)&\(0.24\)&\(3.32\) \\
\bottomrule
\end{tabularx}
\caption{Comparison between Hybrid PT and \sysname on SST-2 and PHEME in accuracy(\%) when using different initial soft prompt strategies. Std. means standard deviation. The best result across different templates is bold and the worst is underlined. }
\label{softpt}
\end{table}

\fakeparagraph{Stability to Soft Prompt Initialization}
We adopt five different soft prompt initialization strategies \cite{gu2022ppt} here to test the stability of our method. The initialization strategies are explained in the Appendix \ref{SPI}.
As shown in \tabref{softpt}, the results indicate that \sysname is more stable to different soft prompt initialization.
The standard deviation of \sysname is lower by 1.19 and 3.08, and the performance of \sysname is better, compared with Hybrid PT on SST-2 and PHEME respectively. 

\fakeparagraph{Stability to Hard Prompt Initialization} We employ ChatGPT \cite{OpenAI2023GPT4TR} to generate multiple expressions with identical meanings (\(T_{sn}\) for SST-2 and \(T_{fn}\) for PHEME) as hard prompts.
The template construction for \(T_{sn}\) and \(T_{fn}\) are provided in the Appendix \ref{HPT}. 
As shown in \tabref{hardpt},  the standard deviation of \sysname is lower by 9.91 and 11.23 compared with Hybrid PT on SST-2 and PHEME, respectively.
The stability of performance indicates the effectiveness of our design which disentangles hard prompt and soft prompt.

\begin{table}[h!]
\centering
\small
\begin{tabularx}{\columnwidth}{p{0.22cm} >{\centering\arraybackslash}X c|p{0.22cm} >{\centering\arraybackslash}X c} 
\toprule
\multicolumn{3}{c|}{\textbf{SST-2}} & \multicolumn{3}{c}{\textbf{PHEME}} \\ 
\midrule
 \(T_{sn}\)   &  \sysname & Hybrid PT  &\(T_{fn}\)     & \sysname & Hybrid PT \\
\midrule
\(T_{s1}\)& $\bm{84.75}$&\(73.98\)&\(T_{f1}\)& $\bm{75.21}$ & $\underline{61.38}$   \\
\(T_{s2}\)& \(83.10\)& \(72.97\) &\(T_{f2}\) & \(72.61\) & \(69.94\)   \\
\( T_{s3} \)& $\underline{81.45 }$& \(69.21\) &\( T_{f3} \)  & \(72.03\)      &\(67.22\)  \\
\( T_{s4} \)& \( 83.44 \)& \(76.21\) &\( T_{f4} \)  & $\underline{70.52} $    &\(67.79\)    \\
\( T_{s5} \)& \(84.23 \)& $\bm{77.63}$ &\( T_{f5} \)  & \(74.64\) & $\bm{73.68}$   \\
\( T_{s6} \)& \(83.95 \)& $\underline{68.63}$& \( T_{f6} \)&\(71.91\)     & \(66.20\)      \\
\midrule
\textbf{Std.} &\(1.11\)&\(11.02\)&\textbf{Std.} &\(2.64\)&\(13.87\) \\
\bottomrule
\end{tabularx}
\caption{Comparison between Hybrid PT and \sysname on SST-2 and PHEME in accuracy(\%) when using different hard prompts. The best result across different templates is bold and the worst is underlined. }
\label{hardpt}
\end{table}

\subsection{Ablation Study}
We conduct an ablation study to analyze the contribution of different modules in \sysname.
For each experiment, we also try 10 random seeds.
The setting of the ablation study is described as follows:

\noindent \textbf{$\textbf{w/o.}$ CL} refers to the model no longer using soft prompts for contrastive learning.
We use only hard prompts and input, processing them through textual information processing and utilizing MLM loss as the total loss.

\noindent \textbf{$\textbf{w/o.}$ GD}  refers to the model lacking a generative decoder.
In this setup, we attach the soft prompt at the beginning of the input and perform contrastive learning on the hidden states of the soft prompt after the semantic encoder's output.


\noindent \textbf{$\textbf{w/o.}$ SP} refers to removing soft prompts and only allowing generative decoder processing contextualized embeddings with both cross entropy and contrastive loss. 

\noindent \textbf{$\textbf{w/o.}$ HP} refers to the model lacking task-specific hard prompts, where we directly attach the [\textit{mask}] token at the beginning of the input.

\begin{table}[h]
    \centering
    \renewcommand\arraystretch{1.2}
    \small
    \begin{tabularx}{\columnwidth}{ p{0.8cm }>{\centering\arraybackslash}X >{\centering\arraybackslash}X  >{\centering\arraybackslash}X >{\centering\arraybackslash}X} 
    \toprule
        \textbf{Method} & \textbf{Gossipop} & \textbf{PHEME} & \textbf{SST-2}  & \textbf{RTE} \\ 
    \midrule
        {StablePT} & $\bm{72.16_{0.72}}$& $\bm{75.35_{0.81}}$ & $\bm{84.04_{0.45}}$ & $\bm{60.34_{0.55}}$  \\  
    \midrule
        {$w/o.$CL} & $69.33_{1.65}$& $73.96_{1.34}$ & $83.11_{0.79}$ & $59.45_{0.91}$  \\ 
        {$w/o.$GD}  & $68.87_{2.15}$ & $69.20_{2.35}$ & $81.19_{2.04}$ & $55.21_{2.23}$  \\
        {$w/o.$SP}  & $68.62_{2.58}$ & $74.31_{1.02}$ & $82.49_{0.94}$ & $54.91_{2.47}$  \\ 
        {$w/o.$HP} & $70.90_{1.23}$ & $73.62_{1.27}$ &$76.87_{1.06}$ &$58.14_{1.45}$  \\ 
    \bottomrule
    \end{tabularx}
    \caption{Ablation study of \sysname  in accuracy (\%).}
    \label{Ablation}
\end{table}
The experiment results shown in \tabref{Ablation} indicate that all four modules contribute to the accuracy improvement. 
For "$w/o.$CL", the overall performance declines, indicating the necessity to acquire class-aware information from instance-level. Moreover, the removal of the generative decoder (\ie "$w/o.$GD") leads to unstable performance, which proves the essence role of this module in stabilizing model performance. For "$w/o.$SP", the overall performance declines, demonstrating soft prompts provide a more direct guide for the model to extract label-specific information from contextualized embeddings. 
"$w/o.$HP" causes a performance drop on all four datasets and the performance on SST-2 suffers the most, which may be attributed to the fact that the sentiment classification task benefits far more knowledge obtained from pre-training than others.
A similar conclusion is also reported in \citet{ni2022electra}.

\subsection{Extension to Larger Model}
To demonstrate that \sysname is equipped with the ability to utilize necessary knowledge from a larger model, we also conduct experiments using Roberta-large as the backbone. We re-trained PPT and SPoT with the same experimental settings as before and the experimental times are shown in Appendix \ref{pretrain_time}.
The experimental results are shown in the Table \ref{PE}.

\begin{table}[H]
\centering
\renewcommand\arraystretch{1.2}
\small
\begin{tabularx}{\columnwidth}{p{0.8cm} >{\centering\arraybackslash}X >{\centering\arraybackslash}X  >{\centering\arraybackslash}X >{\centering\arraybackslash}X} 
\toprule
    {Method}  & \textbf{Gossipop} & \textbf{PHEME} & \textbf{SST-2}  & \textbf{RTE} \\ 
\midrule
    {\sysname} &$\bm{73.32_{0.82}}$& $\bm{77.45_{0.60}}$ & $\bm{88.17_{0.39}}$ & $\bm{62.70_{0.68}}$ \\
\midrule
    {ResPT}&$70.49_{1.98}$ & $73.72_{1.82}$&$78.26_{2.17}$ &$56.79_{2.55}$ \\
    {SPoT}  & $67.45_{1.31}$& $73.38_{1.74}$ & $78.67_{0.51}$ & $53.92_{1.55}$ \\ 
    {PPT}  &$65.21_{1.45}$ & $71.03_{1.91}$ & $78.24_{0.59}$
    &$53.32_{1.42}$ \\ 
    
    {PT} &$63.64_{2.87}$& $70.60_{2.54}$ &$65.84_{4.84}$  &$52.83_{2.53}$  \\ 
    {FT}  &$71.20_{4.41}$&$76.20_{2.29}$  &$85.95_{2.83}$  & $53.54_{1.44}$ \\ 
\bottomrule
\end{tabularx}
    \caption{Parameter efficiency test in  accuracy (\%). PT and FT represent Prompt Tuning and Fine Tuning.}
\label{PE}
\end{table}

It should be noticed that the pre-training times for PPT and SPoT on RoBERTa-large are 8 times and 6 times longer than those on RoBERTa-base respectively. 
The results indicate that our method still maintains a significant advantage over baseline methods, despite the number of parameters in the pre-trained language model (PLM) has increased, demonstrating the universality and efficiency of our model. 
Compared to the results of RoBERTa-base (shown in Table \ref{main_exp}), there is a significant improvement in the effectiveness of all the methods based on RoBERTa-large as expected, which contains more parameters. 
However, it is noteworthy that increasing the number of PLM parameters does not noticeably mitigate the perturbation problems caused by soft prompt initialization, which again emphasizes the importance of stabilizing language model adaptation through the perspective of the model architecture.

\subsection{Extension to Full-data Scenario }
To discuss the performance of \sysname in the full-data scenario, we conduct experiments comparing it with other tuning methods. 
As shown in Table \ref{full}, our method outperforms other prompt-based methods, showcasing its efficiency. 
Additionally, we observe that the standard deviation of the results for all randomly initialized Prompt Tuning methods, including both Prompt Tuning and Residual Prompt Tuning, is more favorable compared to the limited-sample scenario.
This indicates that the perturbations introduced by the random initialization of prompts have been greatly mitigated in the full-data scenario. 
Furthermore, the performance gap between different prompt tuning strategies is notably smaller in the full-data context than in the few-shot scenario. 
This suggests that with sufficient training data, the impact of how prompts are initialized and tuned becomes less pronounced, allowing for more robust and consistent model performance across varying tasks.
\begin{table}[H]
\centering
\renewcommand\arraystretch{1.2}
\small
\begin{tabularx}{\columnwidth}{p{0.8cm} >{\centering\arraybackslash}X >{\centering\arraybackslash}X  >{\centering\arraybackslash}X >{\centering\arraybackslash}X} 
\toprule
    {Method}  & \textbf{Gossipop} & \textbf{PHEME} & \textbf{SST-2}  & \textbf{RTE} \\ 
\midrule
    {\sysname} &$\bm{79.93_{0.77}}$& $\bm{83.65_{0.65}}$ & $\bm{89.43_{0.35}}$ & $\bm{67.38_{0.64}}$ \\
\midrule
    {ResPT}&$79.41_{1.13}$&$82.87_{0.79}$&$89.08_{0.52}$&$65.24_{1.31}$\\
    {SPoT}  & $79.04_{0.82}$& $80.93_{0.86}$ & $85.85_{0.57}$ & $64.43_{0.92}$ \\ 
    {PPT}  &$77.24_{0.69}$ & $78.21_{1.01}$ & $85.32_{0.62}$  & $63.27_{1.14}$ \\  
    {PT} &$76.82_{0.91}$& $77.43_{0.74}$ &$84.51_{0.82}$  &$63.64_{0.89}$  \\ 
    {FT}  &$88.20_{0.45}$&$89.11_{0.38}$  &$94.72_{0.41}$  & $69.52_{0.66}$ \\ 
\bottomrule
\end{tabularx}
    \caption{Full data test in  accuracy (\%).}
\label{full}
\end{table}

\subsection{Extension to Decoder-only PLMs}
To demonstrate the superiority of \sysname on various pre-trained backbones, following the experiment settings of \citet{zhu2023spt}, we run 4 tasks on the GPT-2-small \cite{radford2019language} and LlaMA-2-7b \cite{touvron2023llama} backbones instead of Roberta-base. The results in Appendix \ref{gpt_app} indicate that \sysname works well on the backbones and successfully outperforms the representative prompt tuning and fine-tuning methods.


\subsection{Visualizations}
\begin{figure}[h]
\centering
{\includegraphics[width=0.49\columnwidth]{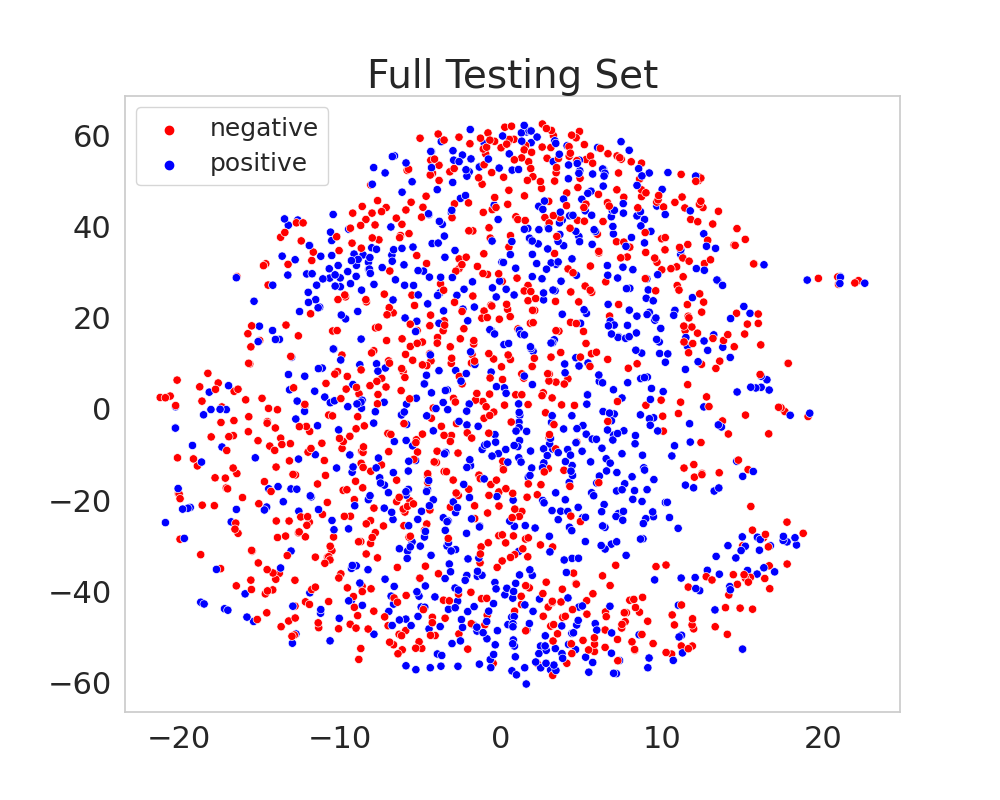}
\includegraphics[width=0.49\columnwidth]{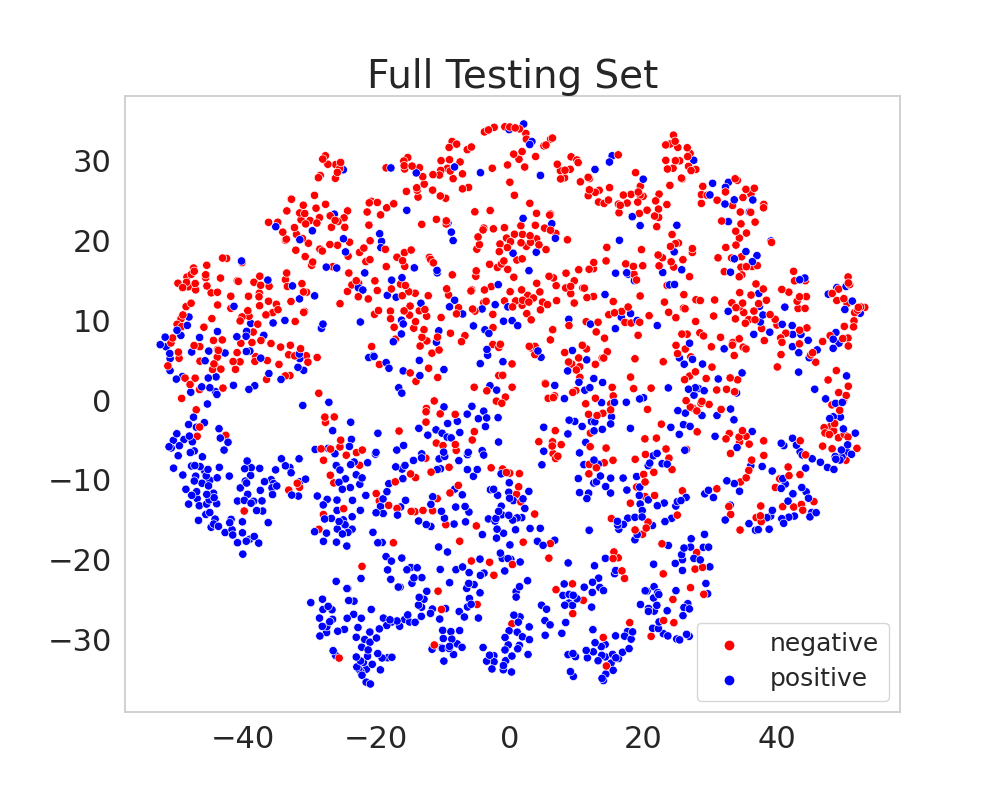}}
\caption{Visualizations of the full testing sets of SST-2 before and after tuning by t-SNE. The left and the right show the results before tuning and after tuning. }
\label{figure_vis}
\end{figure}

To demonstrate that soft prompt processed by the generative decoder exhibits good separability for NLU tasks after training, we plot $\bar{\mathcal{H}_{sp}}$ for the few-shot training and testing data before tuning and after tuning. The underlying dimension reduction and visualization algorithm are t-SNE \cite{van2008visualizing}. The results are illustrated in \figref{figure_vis} (SST-2) and Appendix \ref{visualPHE} (MR, PHEME).
As shown by the results, even reduced in two dimensions, most of the embeddings in the testing
set are clearly separated after tuning, with only a few-shot training samples available on different tasks. 
In addition, the representation of the training sample is widely spread, demonstrating the success in learning class-aware information through contrastive learning and explaining the stability of model performance. 
\begin{table}[h!]
\small
\centering
\begin{tabular}{ccccc}
\toprule
\textbf{Dataset} & \textbf{State} & \textbf{SC$\uparrow$} & \textbf{KL$\uparrow$} & \textbf{MMD$\uparrow$} \\
\midrule
\multirow{2}{*}{SST-2} & before tuning &\(0.01\) &\(0.28\)&\(29.46\) \\
                       & after tuning  &$\bm{0.10}$ &$\bm{0.67}$ &$\bm{379.66}$ \\
\midrule
\multirow{2}{*}{MR}    & before tuning &\(0.001\) &\(0.08\)&\(0.34\) \\
                       & after tuning  &$\bm{0.24} $&$\bm{0.68}$ & $\bm{2295.47}$ \\
\midrule
\multirow{2}{*}{PHEME} & before tuning & \(0.05\)&\(0.12\)&\(200.41\) \\
                       & after tuning  & $\bm{0.18}$&$\bm{0.59}$ & $\bm{1789.42}$\\
\bottomrule
\end{tabular}
\caption{Results of silhouette coefficient (SC), KL divergence (KL), and maximum mean discrepancy (MMD) before and after tuning.}
\label{tab:metric}
\end{table}

To quantitatively assess the intra-class cohesion and inter-class separation of our embeddings, we use three metrics, silhouette coefficient \cite{kullback1951information}, KL divergence \cite{rousseeuw1987silhouettes} and maximum mean discrepancy \cite{gretton2012kernel}. As shown in Table \ref{tab:metric}, the three quantitative metrics all showed a significant increase after training,  further illustrating the effectiveness of \sysname in acquiring category information through soft prompts.

\section{Conclusion}
In this work, we propose \sysname, an innovative hybrid prompt-tuning framework for improving the effectiveness of prompt tuning in few-shot learning and reducing the reliance on well-chosen prompt initialization. 
By introducing a generative decoder module, our system allows soft prompt to interact with the output states of encoded hard prompt and input, rather than simply appending soft prompt to the input text. This approach not only overcomes the initial noise potentially introduced by soft prompts but also enhances the model's ability to capture class-aware information through the application of contrastive learning, thereby achieving better performance in NLU tasks. 
The results across eight datasets in four tasks demonstrate that our method maintains top performance while showing stability to different prompt initializations.

\section*{Limitations}

In this work, we focus on stabilizing model performance across different prompt initializations. However, several limitations still exist for the border applications of \sysname. Firstly, data perturbations still impact our results. Future work could combine data selection~\cite{koksal2022meal,yu2023cold} with \sysname to further enhance performance. Secondly, \sysname does not take the impact of different verbalizers~\cite{cui2022prototypical} into account, we believe that a better verbalizer can further improve language modeling capabilities, as some advanced works have demonstrated the significant impact of verbalizers on performance improvement. Thirdly, \sysname is designed to improve PLMs' performance on NLU tasks, which cannot be directly extended to generation tasks.

\section*{Ethics Statement}
As far as we are aware, our proposed work does not have any ethical considerations. However, our work relies on pre-trained language models, which have been shown to be biased in prior work~\cite{li2021prefix}. As such, users of such models should be aware of and if possible address such issues. The data and the code for this work will be made available to aid reproducibility. Moreover, though all the datasets used in our experiments are publicly available and have not been reported to carry a social bias against any sensitive attributes, and the proposed approach would not explicitly introduce new negative societal impacts, more work is still needed to investigate the potential unfairness in these datasets.

\section*{Acknowledgements}
We thank all the anonymous reviewers and the area chair for their helpful feedback, which aided us in greatly improving the paper.
This work is supported by National Natural Science Foundation of China (62272371, 62103323, U21B2018, 62161160337,  U20B2049), Initiative Postdocs Supporting Program (BX20190275, BX20200270), China Postdoctoral Science Foundation (2019M663723, 2021M692565), Fundamental Research Funds for the Central Universities under grant (xzy012024144), and Shaanxi Province Key Industry Innovation Program (2021ZDLGY01-02).

\bibliography{anthology,custom}
\bibliographystyle{acl_natbib}

\clearpage
\appendix

\section{Experiment Details} \label{DatasetInfo}
We choose Roberta-base~\cite{liu2019roberta} as our backbone model.
For all tasks, we follow the same procedure as \citet{gu2022ppt} to form the true few-shot learning settings \cite{perez2021true}. 
In particular, we randomly select 64 samples from the original training set to construct a few-shot training set, and construct a development set by randomly selecting another 64 samples from the original training set.
We ensure that the number of labels is balanced for both training and development sets. 
We apply the AdamW~\cite{adamw} optimizer with a learning rate of 1e-4 and weight decay of 1e-4, and train for 100 epochs. The mini-batch size is 8. For all prompt-based methods, we set the prompt lengths as 10. 
We run experiments with 10 different random seeds on a single NVIDIA A100 and report the average accuracy and standard deviation.

Giving details of the datasets used in our main experiments
in the few-shot setting, including type, label words, and size of test in Table \ref{datasetinfo}. \(|C|\): The number of categories for tasks.

\begin{table}[h]
\renewcommand\arraystretch{1.2}
\centering
\small
\begin{tabularx}{\columnwidth}{p{1.4cm}|c c c c}
\toprule
\textbf{Datasets} &Type &\(|C|\) &\(|Test|\) & Label words\\
\midrule
\textbf{Politifact}&FD& 2&19k &fake/real  \\
\textbf{Gossipop}&FD& 2 & 32k & fake/real \\
\textbf{PHEME}&FD& 2 &   1.6k & fake/real \\
\textbf{SST-2}&SA& 2&    1.8k     &positive/negative \\
\textbf{MR}  &SA & 2   & 2k&positive/negative  \\
\textbf{QNLI}&NLI& 2 &   5.3k         &  yes/no  \\
\textbf{RTE} &NLI&2  &   2.4k         & Clearly/Yet\\
\textbf{BoolQ} & QA&2  & 3.2k         & yes/no\\
\bottomrule
\end{tabularx}
\caption{Datasets information in the main experiments. FD means fake news detection, SA means sentiment analysis, NLI means natural language inference and QA means question answer. }
\label{datasetinfo}
\end{table}

\section{Pseudocode of \sysname}
Pseudocode of \sysname is presented in Algorithm \ref{alg:stbpt}.

\label{pscode}
\begin{algorithm}[t]
\caption{Algorithm of StablePT}
\begin{algorithmic}[1]
\Require{Input $x$, hard prompt $p_{h}$, soft prompt $p_{s}$ batch size $b$ and labels $y$}
\Ensure{A learned model StablePT, consisting of semantic encoder $f_{s}$ with parameters $\theta_{S}$, generative decoder $f_{g}$ with parameters $\theta_{G}$, soft prompt $p_{s}$ with parameters $\theta_{s}$, PLM Encoder with parameters $\theta_{P}$} 
\State Random initialize $\theta_{S}$, $\theta_{G}$, $\theta_{s}$
\State Freeze $\theta_{P}$
\State $epoch \gets 0$
\While{ $epoch \leq epoch_{max}$}
    \State $n \gets 0$
    \While{ $n \leq n_{max}$}
       \State Randomly select batch $b_{n}$
       \State $\mathcal{H}_{se}=f_{s}(p_{h},b_{n})$
       \State $\mathcal{H}_{sp}=f_{g}(p_{s},\mathcal{H}_{se})$
       \State Calculate $\mathcal{L}_{CL}$ with equation \eqref{eq5}, calculate $\mathcal{L}_{MLM}$ with equation \eqref{eq6}, calculate $\mathcal{L}_{total}$ with equation \eqref{eq7}
       \State Backward on $\mathcal{L}_{total}$ and update $\theta_{S}$, $\theta_{G}$, $\theta_{s}$ based on AdamW gradient descent with an adjustable learning rate
        \State $n \gets n+1$
       \EndWhile
    \State $epoch \gets epoch+1$
\EndWhile \\
\Return A trained model StablePT
\end{algorithmic}\label{alg:stbpt}
\end{algorithm}

\begin{figure*}[t]
\centering
\begin{subfigure}[b]{0.34\columnwidth}
\includegraphics[width=\linewidth]{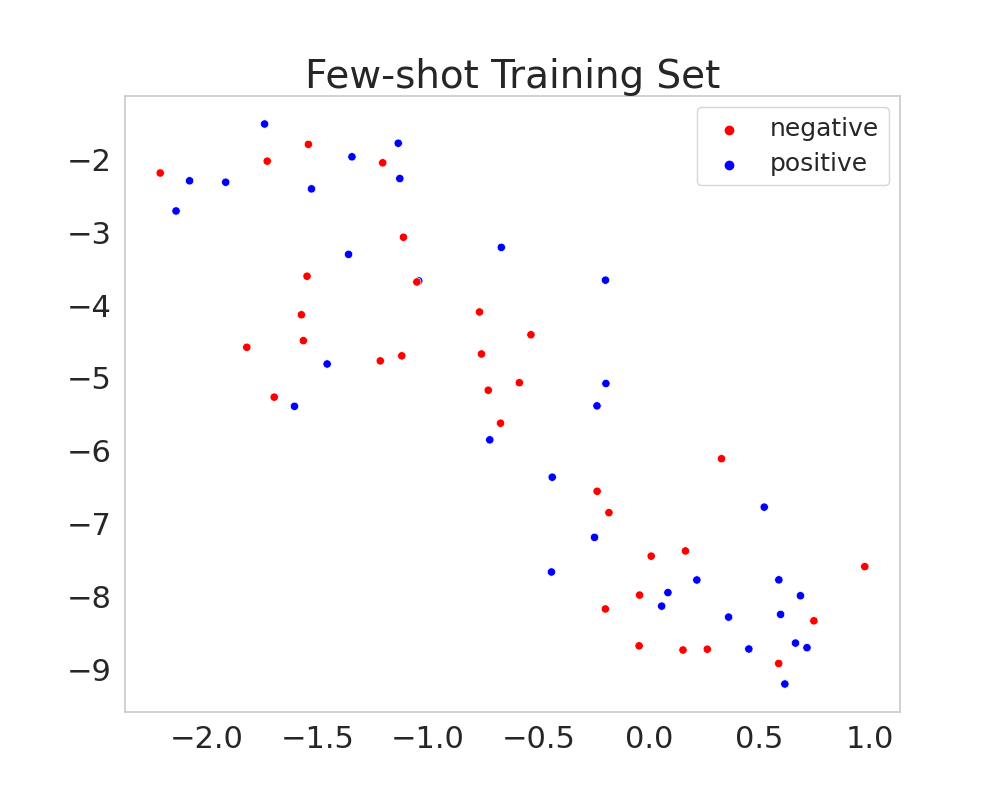}
\end{subfigure}
\begin{subfigure}[b]{0.34\columnwidth}
\includegraphics[width=\linewidth]{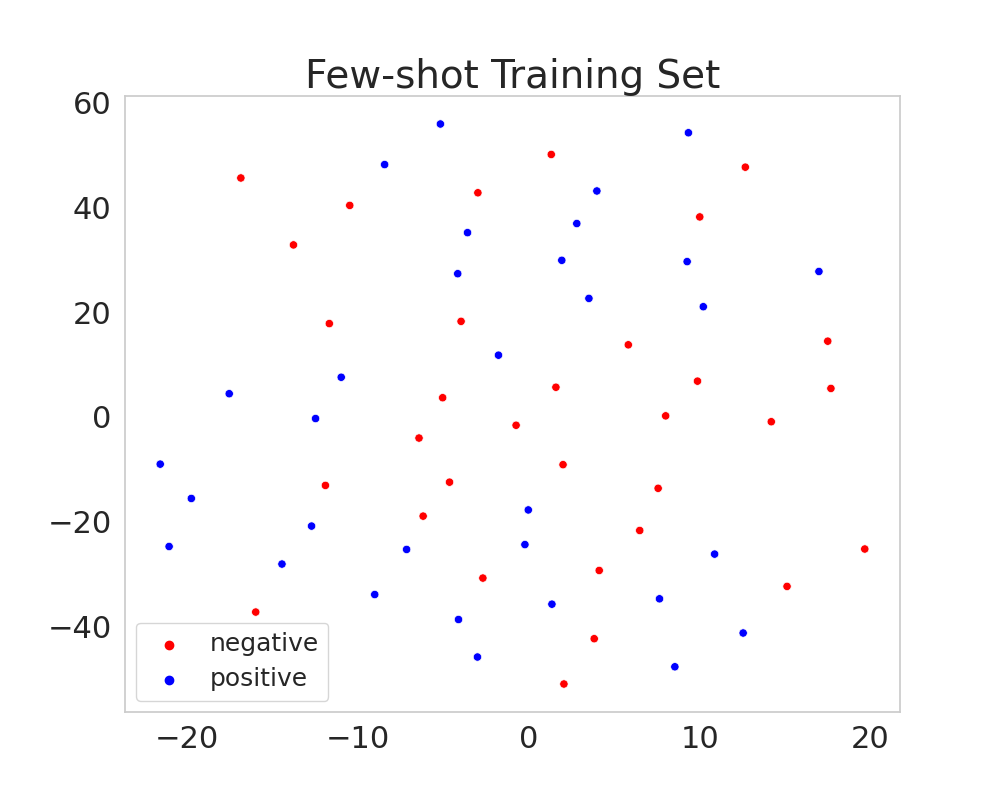}
\end{subfigure}
\begin{subfigure}[b]{0.34\columnwidth}
\includegraphics[width=\linewidth]{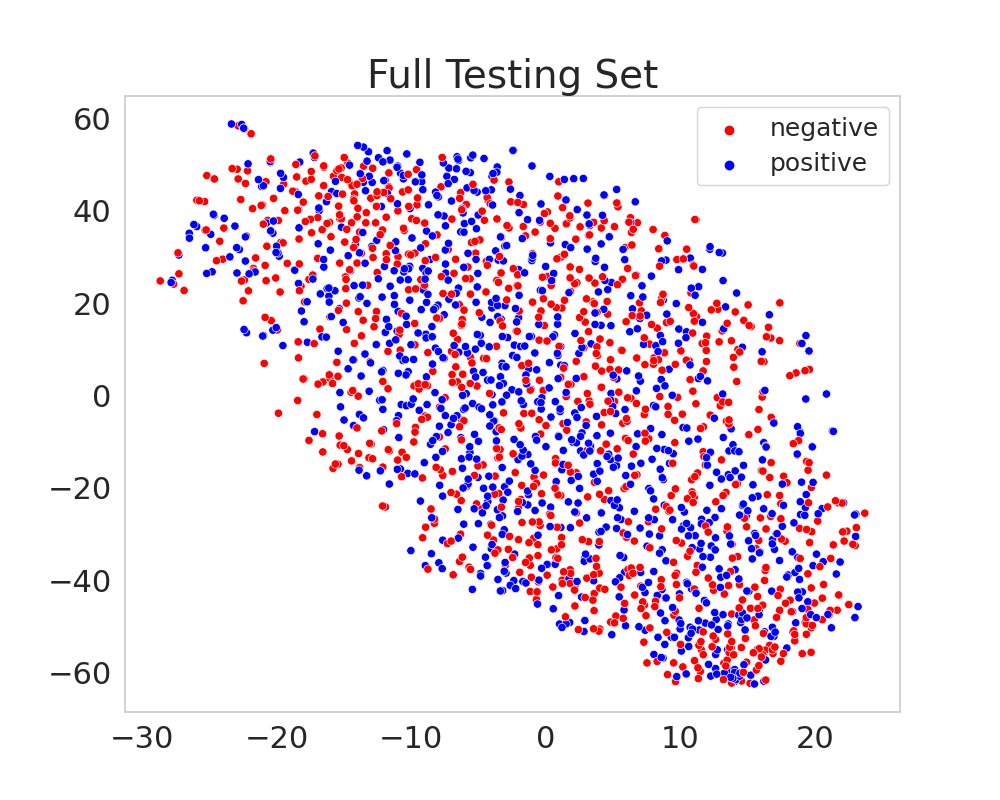}
\end{subfigure}
\begin{subfigure}[b]{0.34\columnwidth}
\includegraphics[width=\linewidth]{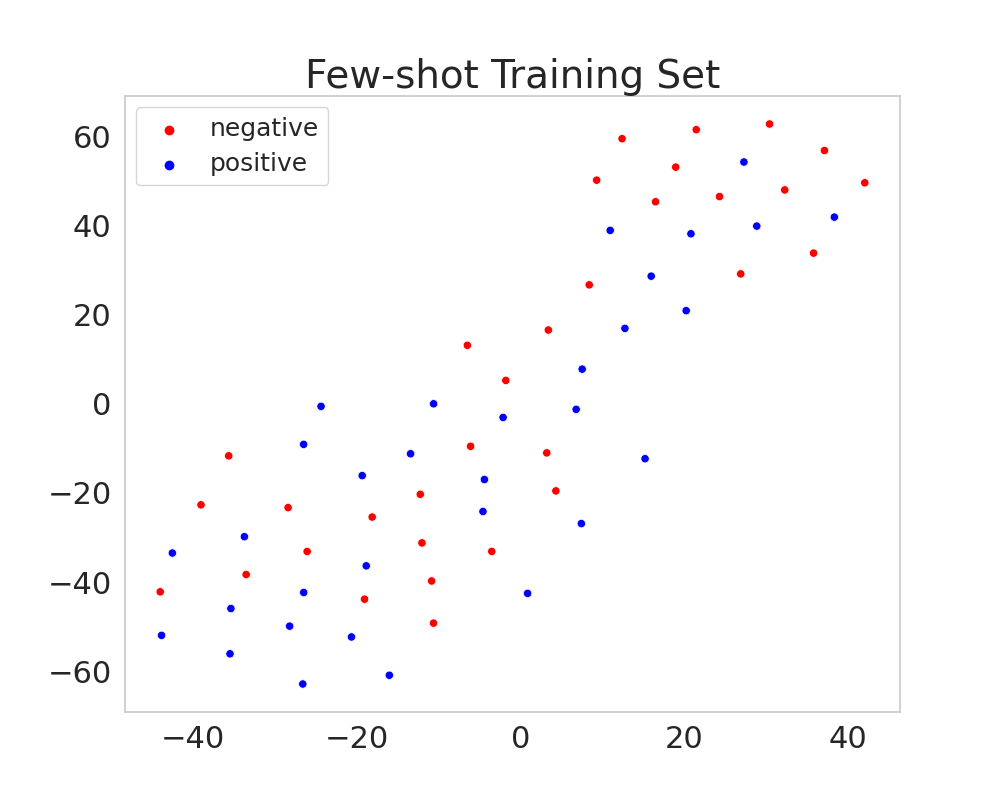}
\end{subfigure}
\begin{subfigure}[b]{0.34\columnwidth}
\includegraphics[width=\linewidth]{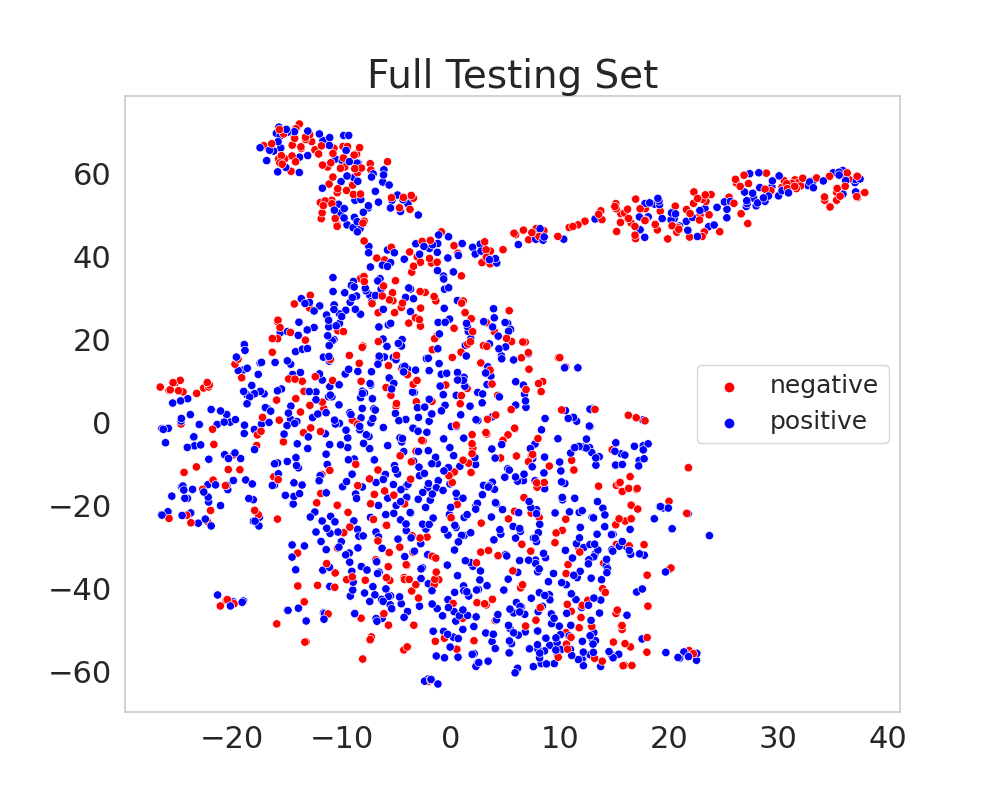}
\end{subfigure}
\begin{subfigure}[b]{0.34\columnwidth}
\includegraphics[width=\linewidth]{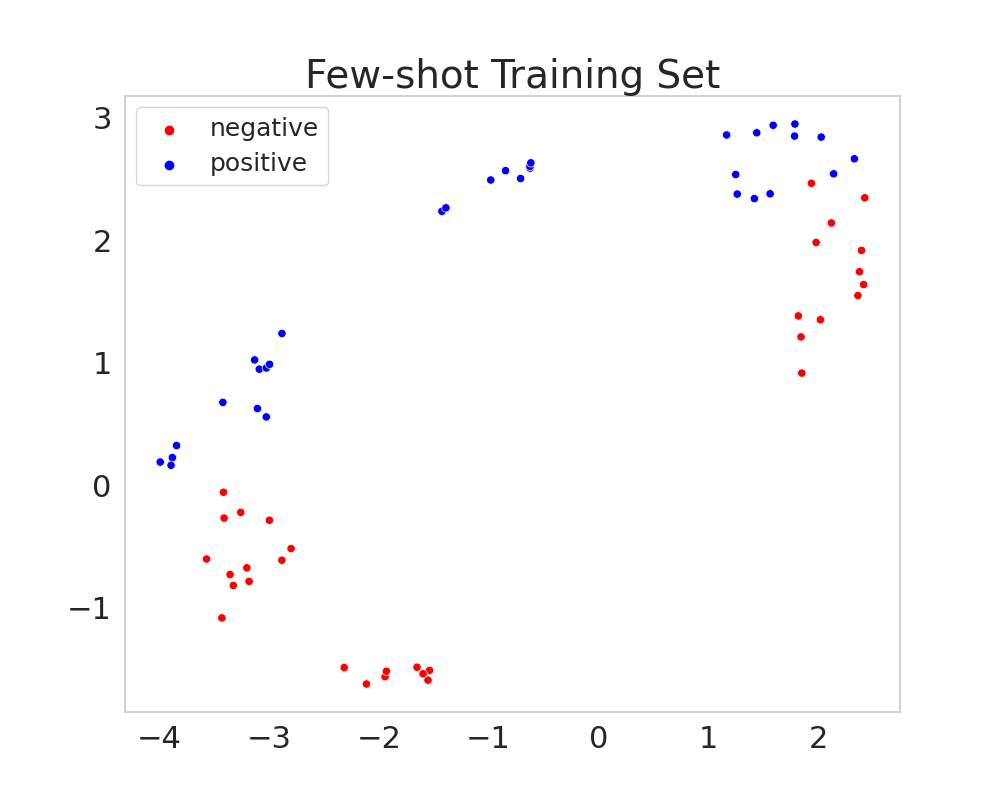}
\subcaption{Dataset: SST-2}
\end{subfigure}
\begin{subfigure}[b]{0.70\columnwidth} 
    \includegraphics[width=0.49\linewidth]{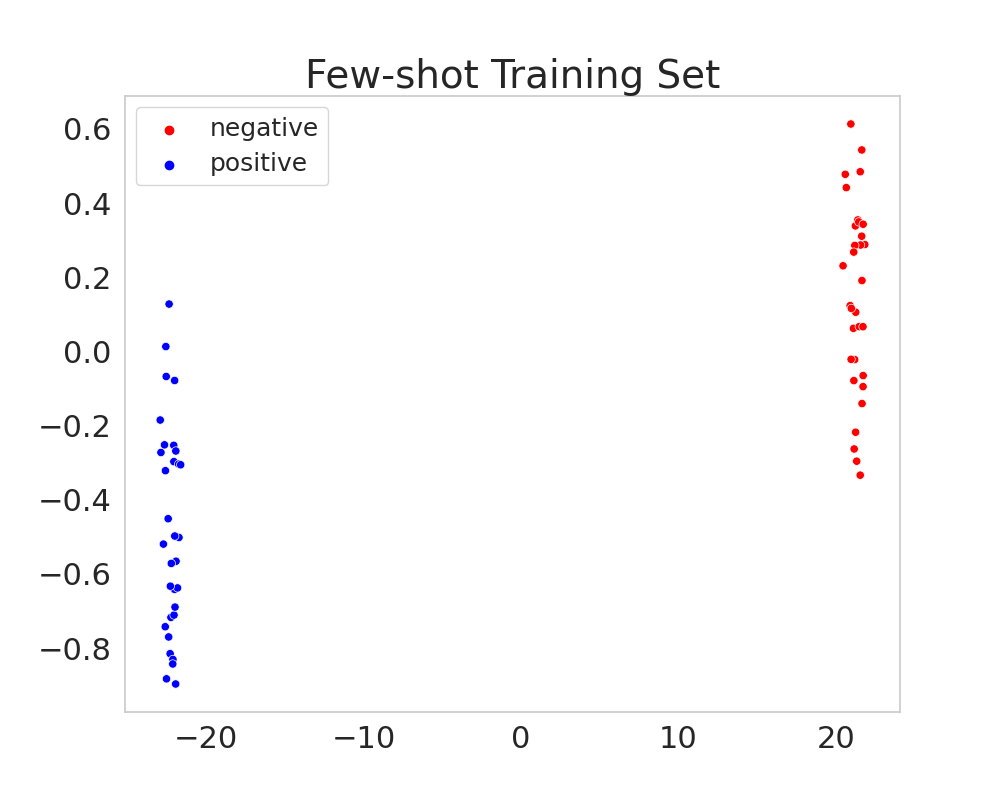}
    \includegraphics[width=0.49\linewidth]{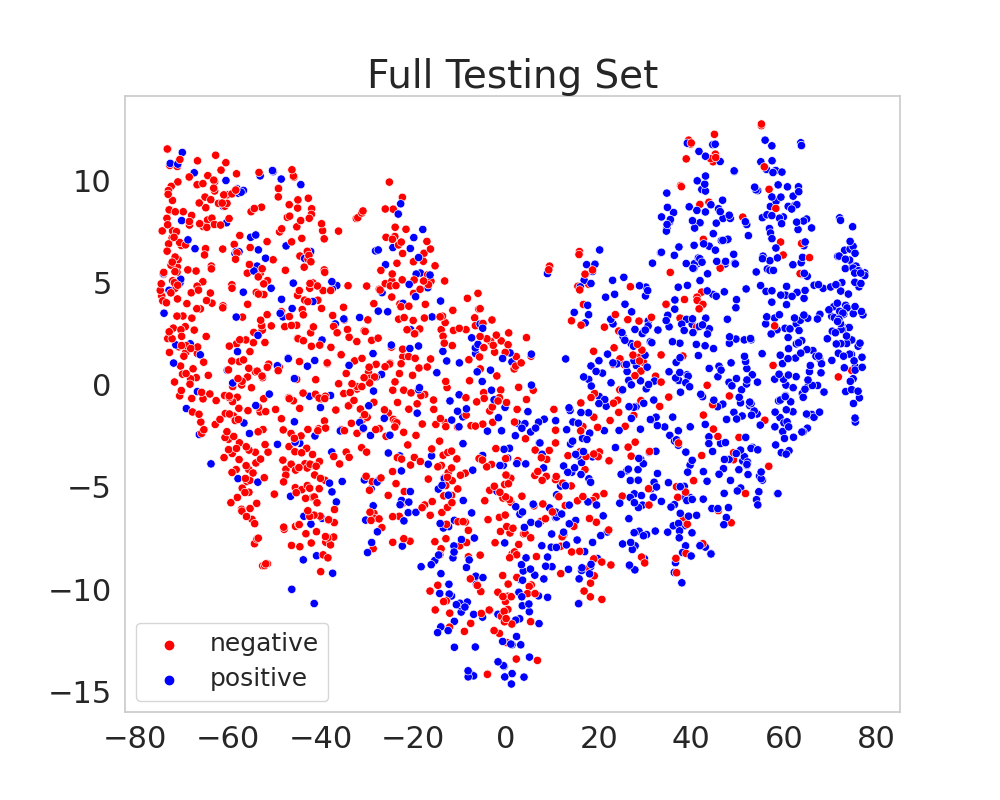}
    \subcaption{Dataset: MR}
\end{subfigure}
\begin{subfigure}[b]{0.70\columnwidth} 
    \includegraphics[width=0.49\linewidth]{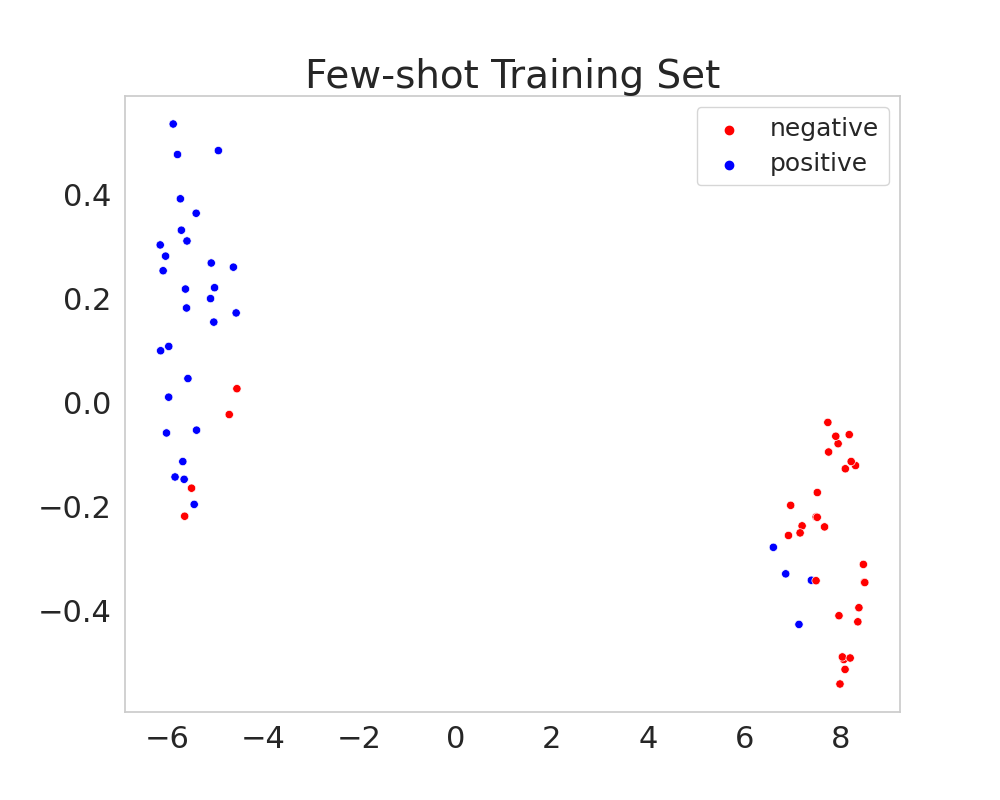}
    \includegraphics[width=0.49\linewidth]{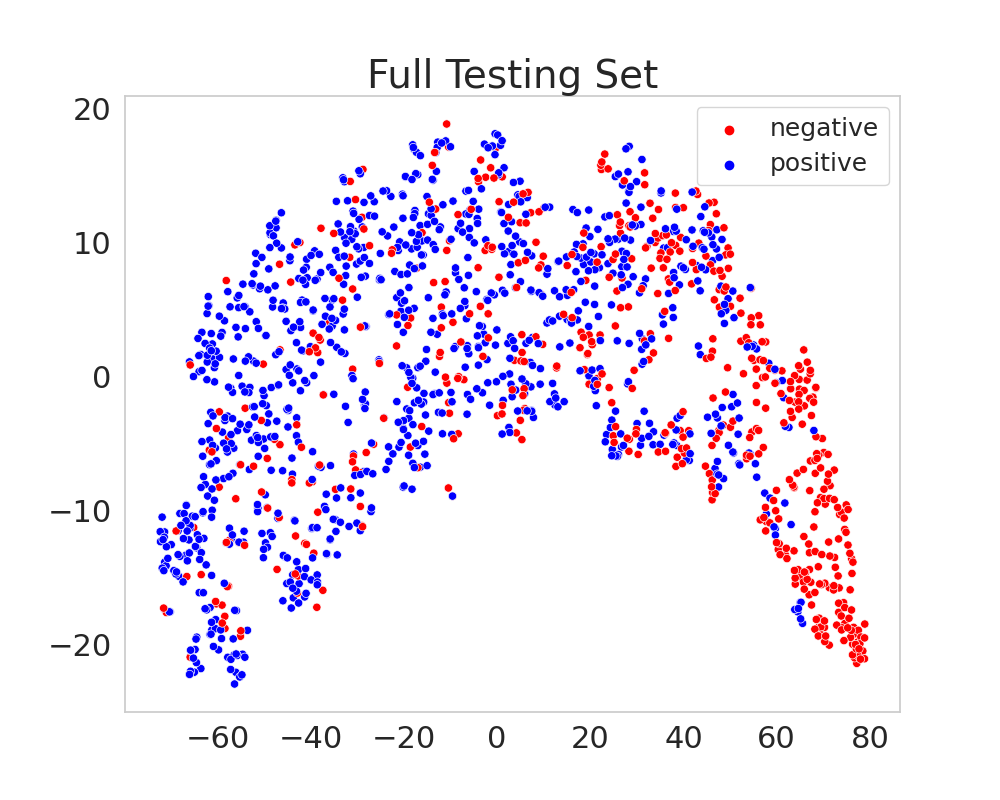}
    \subcaption{Dataset: PHEME}
\end{subfigure}
\caption{Visualization of the few-shot training sets and the full testing sets of SST-2, MR, and PHEME by t-SNE. The top shows the results before tuning and the bottom shows the results after tuning.}
\label{fig:pheme}
\end{figure*}

\section{Details of Comparison Methods}\label{methods}
In  \tabref{tab:tuning_methods}, we introduce specific
details of \sysname and other comparison methods.
\begin{table}[t]
\centering
\begin{tabular}{l c c}
\toprule
Method        & Train. params & Pre-train \\
\midrule
Prompt Tuning    & 7.7K   & No         \\
Prefix Tuning  & 14.2M   & No         \\
P-TuningV2     & 14.2M      & No       \\
PPT             & 7.7K     & Yes      \\
SPoT          & 7.7K       & Yes      \\
ResidualPT& 9.3M    & No            \\
SMoP             & 77K        & No     \\
E$^{2}$VPT             & 46K        & No    \\
Fine Tuning      & 135M       & No     \\
CP-Tuning        & 135M       & No   \\
Stable PT    & 13.1M      & No        \\ \bottomrule
\end{tabular}
\caption{Comparison of all methods. Train. params denotes total number of trainable
parameters, Pre-train denotes if the method
requires pre-training on source tasks.}
\label{tab:tuning_methods}
\end{table}

\section{Soft Prompt Initial Strategies}\label{SPI}
In \secref{HPI}, we utilized five methods of soft prompt initialization, and here we provide specific explanations for these five methods.
"Random" indicates that we randomly initialize the embedding of soft prompts. 
"Label" indicates that we use the embeddings of the label words. 
"Vocab" indicates that we randomly sample words from the vocabulary.
"Top-1k" indicates that we randomly sample words from the most frequent 1000 words in the pre-training corpus. 
"Task" indicates that we randomly sample words from the downstream data.

\section{Hard Prompt Template}\label{HPT}
 Table~\ref{tab:my_label_senti} and Table~\ref{tab:my_label_fake} show the hard prompt template used in \secref{HPI} for sentiment analysis and fake news detection, respectively. All these prompts are generated by employing ChatGPT \cite{OpenAI2023GPT4TR} to express the same meaning but with different vocabularies.
    \begin{table}[h!]
    \renewcommand\arraystretch{1.2}
    \centering
    \small
    \begin{tabularx}{\columnwidth}{p{0.2cm} c}
    \toprule
    $\bm{T_{sn}}$   & $\bm{Template}$ \\
    \midrule
    $T_{s1}$& \textit{"The sentiment of the review is <mask> ."}  \\
    $T_{s2}$& \textit{"The outlook portrayed in this appraisal is <mask> ."} \\
    $ T_{s3}$& \textit{"The emotional tone of this testimonial is <mask> ."} \\
    $ T_{s4}$& \textit{"The sentiment of this critique is <mask> ."} \\
    $T_{s5}$&\textit{"The feeling conveyed by this evaluation is <mask> ."}\\
    $T_{s6} $&\textit{"The mood expressed in this assessment is <mask> ."} \\
    \bottomrule
    \end{tabularx}
    \caption{Example of hard prompt template for sentiment analysis}
    \label{tab:my_label_senti}
    \end{table}
    
    \begin{table}[h!]
    \renewcommand\arraystretch{1.2}
    \centering
    \small
    \begin{tabularx}{\columnwidth}{p{0.2cm} c}
    \toprule
    $\bm{T_{fn} }$   & $\bm{Template}$ \\
    \midrule
    $T_{f1}$& \textit{"Here is a piece of claim with <mask> information ."}  \\
    $T_{f2}$& \textit{"The outlook portrayed in this appraisal is <mask> ."} \\
    $T_{f3}$& \textit{"The content of this statement is <mask> ."} \\
    $ T_{f4} $& \textit{"What this declaration says is <mask> ."} \\
    $T_{f5} $&\textit{"The essence of this proclamation is <mask> ."}\\
    $ T_{f6}$&\textit{"This announcement conveys a <mask> message ."} \\
    \bottomrule
    \end{tabularx}
    \caption{Example of hard prompt template for fake news detection}
    \label{tab:my_label_fake}
    \end{table}

\section{Visualizations} \label{visualPHE}
We plot the average-pooled soft prompt (\ie $\bar{\mathcal{H}_{sp}}$) of the few-shot training and testing data in MR and PHEME, shown in \figref{fig:pheme}.

\section{Impact of Hyperparameters}
\label{hyperparameter}
\subsection{Sample Efficiency}
\begin{figure}[h]
\centering
\begin{subfigure}[b]{0.48\columnwidth}
    \centering
    \includegraphics[width=\linewidth]{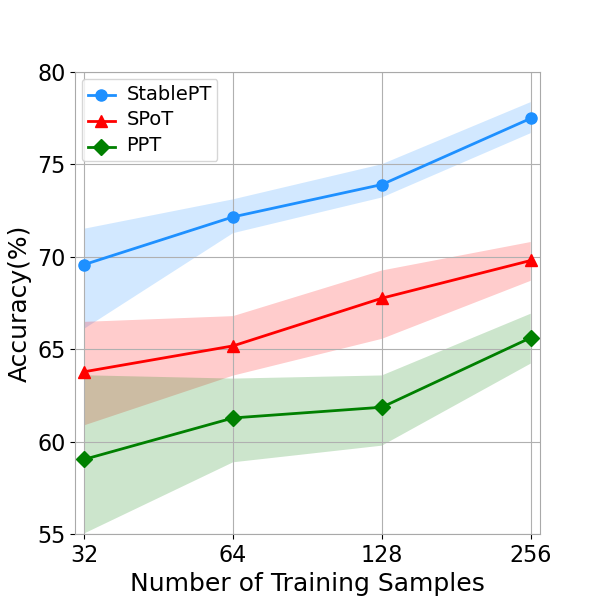}
    \caption{Gossipop}
\end{subfigure}
\hfill
\begin{subfigure}[b]{0.48\columnwidth}
    \centering
    \includegraphics[width=\linewidth]{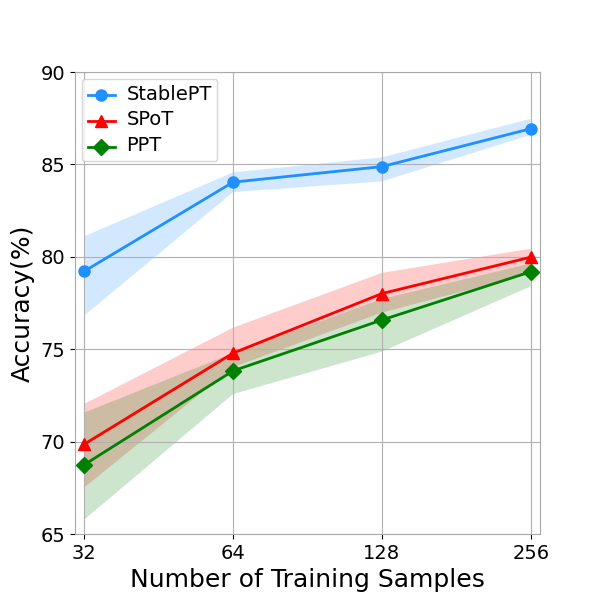}
    \caption{SST-2}
\end{subfigure}
\caption{Comparison among PPT, SPoT and \sysname with different numbers of training samples on Gossipop and SST-2.}
\label{fig:se}
\end{figure}

We'd like to discuss how performance varies with an increasing number of training samples.
We select two strong baselines SPoT and PPT for comparison.
As shown in the \figref{fig:se}, with the number of training samples increasing, \sysname, PPT, and SPoT all exhibit a gradual upward trend in results.
Besides, \sysname outperforms PPT and SPoT across all scales of training samples for different tasks with slight fluctuations. In addition, the suboptimal performance caused by the domain inconsistency between pre-training and downstream tasks (mentioned in \secref{sec:main}) still exists with the increase in the number of training samples.

\subsection{Prompt Length}
The length of the soft prompt could act as an important hyperparameter.
Therefore, we investigate the impact of various lengths of soft prompt on \sysname across four different datasets, as shown in \tabref{pmptlength}.

\begin{table}[h]
    \centering
    \renewcommand\arraystretch{1.2}
    \small
    \begin{tabular}{lcccc} 
    \toprule
        \textbf{PL} & \textbf{Gossipop} & \textbf{PHEME} & \textbf{SST-2}  & \textbf{RTE} \\ 
    \midrule
        {5} & ${71.17_{1.04}}$&${74.61_{1.23}}$ & ${83.29_{1.43}}$ & ${59.21_{1.03}}$  \\ 
        {10} & $72.16_{0.72}$&$75.35_{0.81}$  &${84.04_{0.45}}$& ${60.34_{0.55}}$ \\ 
        {20} &$70.66_{2.02}$&$73.53_{1.88}$ &$83.12_{1.79}$ &$58.83_{2.23}$  \\ 
        {50} &$70.53_{1.55}$&$73.24_{1.29}$&$82.99_{1.08}$& $59.34_{1.81}$ \\ 
    \bottomrule
    \end{tabular}
    \caption{\sysname performance with various soft prompt lengths (PL) in accuracy (\%).}
    \label{pmptlength}
\end{table}
The results, which explore soft prompt lengths of 5, 10, 20, and 50, reveal a discernible pattern, \ie increasing the prompt length beyond 10 does not proportionally enhance performance. 
In fact, longer prompts often lead to a marginal decline in effectiveness, possibly due to dilution of semantic density or increased complexity in the learning process \cite{lester-etal-2021-power}. Choosing an optimal length can facilitate PLMs to extract subtle semantic insights without being burdened by redundant information. This underscores the importance of prompt length optimization in maximizing the efficacy of soft prompts in contrastive learning settings.
Especially when balancing the need for contextual richness against the constraints of computational efficiency and model generalizability.

\section{Extension to Decoder-only PLMs}\label{gpt_app}
The experimental results of decoder-only backbones are shown in Table \ref{gpt}.

\begin{table}[h]
\centering
\renewcommand\arraystretch{1.2}
\small
\begin{tabularx}{\columnwidth}{p{0.8cm} >{\centering\arraybackslash}X >{\centering\arraybackslash}X  >{\centering\arraybackslash}X >{\centering\arraybackslash}X} 
\toprule
    {Method}  & \textbf{Gossipop} & \textbf{PHEME} & \textbf{SST-2}  & \textbf{RTE} \\ 
\midrule
    \multicolumn{5}{c}{\textbf{GPT-2 backbone}}\\
\midrule
    {\sysname} &$\bm{67.78_{0.46}}$& $\bm{73.33_{0.42}}$ & $\bm{74.83_{0.67}}$ & $\bm{52.55_{0.34}}$ \\
\midrule
    {ResPT}&$64.26_{4.53}$ & $72.41_{1.78}$&$63.86_{3.63}$ &$51.10_{0.57}$ \\
    {PT} &$60.15_{2.81}$& $69.40_{3.11}$ &$60.52_{3.63}$  &$50.34_{0.63}$  \\ 
    {FT}  &$67.21_{2.63}$&$73.02_{2.62}$  &$66.21_{2.63}$  & $51.48_{1.03}$ \\ 
\midrule
    \multicolumn{5}{c}{\textbf{LlaMA-2 backbone}} \\
\midrule
    {\sysname} &$\bm{71.13_{0.63}}$& $\bm{75.59_{0.73}}$ & $\bm{81.93_{0.10}}$ & $\bm{52.78_{0.36}}$ \\
\midrule
    {ResPT}&$67.56_{3.64}$ & $73.52_{2.07}$&$66.90_{3.98}$ &$51.46_{0.90}$ \\
    {PT} &$63.57_{1.87}$& $68.92_{1.14}$ &$62.63_{3.84}$  &$51.03_{0.84}$  \\ 
    {FT}  &$68.21_{2.07}$ &$75.04_{1.61}$  &$81.25_{1.24}$  & $52.48_{1.83}$ \\ 
\bottomrule
\end{tabularx}
    \caption{Various PLM backbones tests in accuracy (\%).}
\label{gpt}
\end{table}
The results indicate that our method still maintains a significant advantage over baseline methods, despite using decoder models of different sizes as the backbones, demonstrating the universality of our model.

\section{Time Using of Pretrain Methods}\label{pretrain_time}

For a fair comparison, we use the datasets from the original paper to re-train PPT and SPoT on Roberta-base and Roberta-large. Table \ref{pretrain} shows the time required for the experiment.

\begin{table}[h]
    \renewcommand\arraystretch{1.2}
    \centering
    \small
    \newcolumntype{C}{>{\centering\arraybackslash}X}
    \begin{tabularx}{\columnwidth}{p{1.5cm} C C}
        \toprule
        \textbf{Methods} & \textbf{Roberta-base} & \textbf{Roberta-large} \\
        \midrule
        \textbf{PPT} & 93123s & 781934s \\
        \textbf{SPoT} & 110954s & 676522s \\
        \bottomrule
    \end{tabularx}
    \caption{Consumption of PPT and SPoT in time (s).}
    \label{pretrain}
\end{table}
The results indicate that as the model parameters increase, the pretraining time required for PPT and SPoT also increases, which consume substantial computational resources. This may hinder the further development of pretraining prompt methods.

\end{document}